\documentclass[english]{lni}
\usepackage[latin1]{inputenc}
\usepackage{hyperref}
\usepackage{graphicx}
\usepackage{fancyhdr}
\usepackage{listings} %if lstlistings is used
\usepackage{changepage} %for changing topmargin on first page
\usepackage[figurename=Fig., tablename=Tab., small]{caption}[2008/04/01]

\usepackage[caption=false]{subfig}
\usepackage{multirow}
\fancypagestyle{titlepage}{
\fancyfoot{}}
\renewcommand{\headrulewidth}{0.4pt} %horizontal line below header
\author{Alperen Kantarc{\i}\footnote{Department of Computer Engineering, Istanbul Technical University, Turkey, \{kantarcia,ekenel\}@itu.edu.tr}   , Hasan Dertli\footnote{Sodec Technologies, Istanbul, Turkey, hasan.dertli@sodecapps.com}  , Haz{\i}m Kemal Ekenel\footnotemark[1]}
\title{\vspace{12pt} Shuffled Patch-Wise Supervision for Presentation Attack Detection}
\begin{document}

\maketitle

\renewcommand{\refname}{References}
\setcounter{footnote}{2} %Change to the number of authors for a correct numbering of the foot notes
\thispagestyle{titlepage}
%header setting after the second page
\pagestyle{fancy}
\fancyhead{} % clears header settings
\fancyhead[RO]{\small Shuffled Patch-Wise Supervision for Presentation Attack Detection \hspace{25pt} \hspace{0.05cm} }
\fancyhead[LE]{\hspace{0.05cm}\small \hspace{25pt} Alperen Kantarc{\i}, Hasan Dertli, Haz{\i}m Kemal Ekenel}
%\fancyhead[LE]{\hspace{0.05cm}\small \thepage \hspace{5pt} Anonymous submission}
\fancyfoot{} % clears all footer settings
\renewcommand{\headrulewidth}{0.4pt} %line below header

\begin{abstract}
Face anti-spoofing is essential to prevent false facial verification by using a photo, video, mask, or a different substitute for an authorized person's face. Most of the state-of-the-art presentation attack detection (PAD) systems suffer from overfitting, where they achieve near-perfect scores on a single dataset but fail on a different dataset with more realistic data. This problem drives researchers to develop models that perform well under real-world conditions. This is an especially challenging problem for frame-based presentation attack detection systems that use convolutional neural networks (CNN). To this end, we propose a new PAD approach, which combines pixel-wise binary supervision with patch-based CNN. We believe that training a CNN with face patches allows the model to distinguish spoofs without learning background or dataset-specific traces. We tested the proposed method both on the standard benchmark datasets ---Replay-Mobile, OULU-NPU--- and on a real-world dataset. The proposed approach shows its superiority on challenging experimental setups. Namely, it achieves higher performance on OULU-NPU protocol 3, 4 and on inter-dataset real-world experiments.%we propose a new training method for frame-based presentation attack detection systems to increase the robustness of the system against real-world conditions. 
%We train a previously proposed model, DeepPixBis, with the proposed training method and benchmarked it both on the standard datasets, as well as on a real-world dataset. 

\end{abstract}

\section{Introduction}
%Face anti-spoofing has been a popular research area for various security applications~\cite{fas_china_applications}. 
In recent years, facial recognition systems are widely used as they are robust and reliable for common usage. However, these recognition systems have to be careful about the authenticity of a given face input. If the given input is recorded from a video of an authorized user, the recognition system should not recognize the person in the video and give access to the system. Presentation attack detection (PAD) systems aim to prevent this problem by evaluating the liveness of the given person's image. 

In recent years, PAD methods improved significantly with the progress in deep learning methods and publicly available large, representative datasets \cite{OULU_NPU_2017,siwdataset,casiasurf,replaymobile}. Most of the significant progress has been achieved when researchers found different cues to decide liveness of a face \cite{rrpg2016,lbpbased,patch_depth_cnn}.  These different cues used with complex deep neural networks to create PAD systems that are very successful in intra-dataset benchmark results. However, the real challenge in PAD still remains as an inter-dataset benchmark which shows the real performance of the PAD systems in real-world like scenario. Most of the systems that use CNNs overfit the data easily by memorizing reflection and illumination effects. To address this problem, in this paper, we propose a new training procedure for face PAD systems. We show that our training method utilizes the pixel-wise binary loss in a better way. Moreover, we show that our proposed method improves model performances on real-world experiments. %cross-dataset experiments which are the challenging and most important benchmark for the PAD systems. 

%The contributions of this paper can be summarized as follows: First, we propose a new training procedure for face PAD systems. Second, we show that our training method utilize the pixel-wise binary loss better. Finally,  we show that our proposed method improve model performances on cross-dataset experiments which are the challenging and most important benchmark for the PAD systems. 

%The remainder of this paper is organized as follows. In Section 2, we give a brief overview of the related work. Then, in Section 3, we explain the utilized model and the approach. In Section 4, details of the datasets, benchmarks, and experimental results are presented and discussed. The last section concludes the paper and gives directions for the future work.

\begin{figure}[!t]
	\centering
	\includegraphics[width=0.8\linewidth]{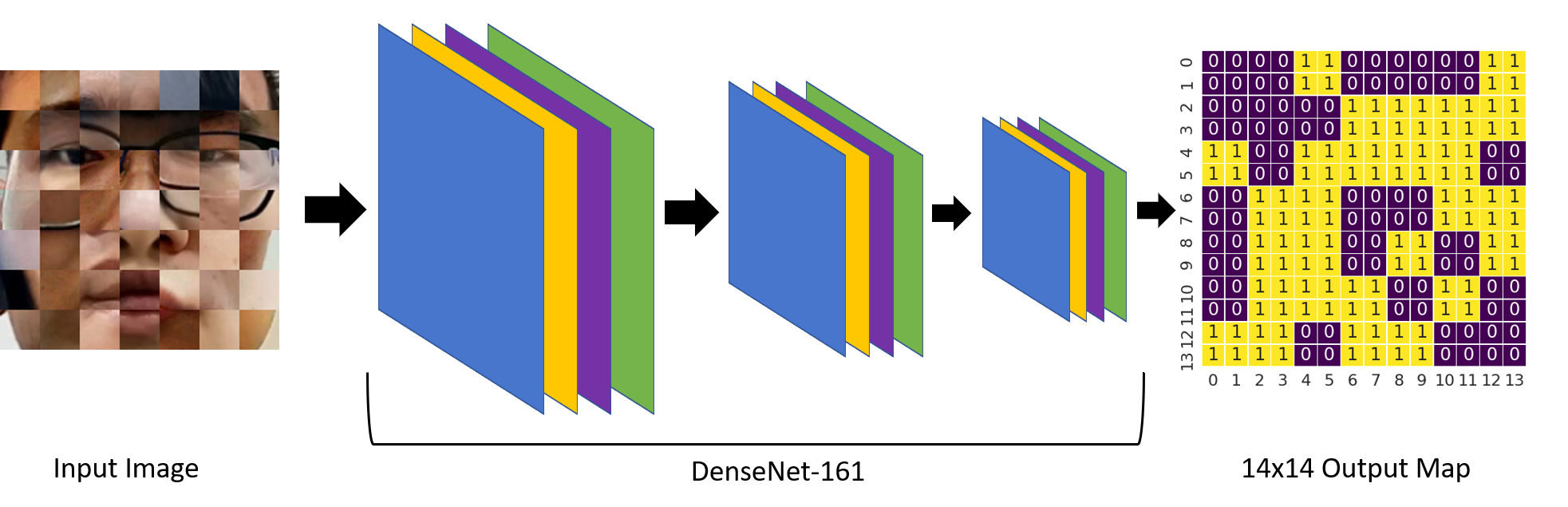}
	\caption{\label{fig:arch-illustration}Overview of the proposed method}
\end{figure}

\section{Related Work}
PAD approaches are mainly grouped into two categories; video-based and frame level. While video-based methods use temporal consistency and temporal cues, frame-based methods use subtle cues from the given face images. These cues can be summarized as liveness, texture, and 3D geometry \cite{2010survey}. Among these cue-based methods, liveness cues are applicable for video-based PAD. Therefore, texture and 3D geometry cue-based methods are more popular for detecting liveness from a frame. 3D geometry cue-based methods either use depth or pseudo-depth signals to distinguish attacks from real attempts. Even though devices with depth sensors, time-of-flight cameras, Lidar sensors, etc., are getting popular in daily usage, most mobile phones or video cameras do not have depth sensors. Therefore, methods mostly rely on pseudo-depth maps which are not real data and may not reflect real-world data distribution very well. Therefore, most of these methods might get good results on specific datasets but fail to generalize. %We briefly present frame-based methods that use texture or pseudo-depth map cues.

As initial work with deep neural networks, \cite{yang2014learn} proposes to use a face alignment network for preparing face images to train an AlexNet~\cite{alexnet} model. They use the model for extracting features of the face and use an SVM classifier to classify images as artefact or bona fide.

In order to improve PAD performances, researchers search different supervisions, along with the binary classification objective, for training their models. \cite{deeppixbis} proposes an effective model for frame-level PAD.  They add additional supervision, which they call pixel-wise binary supervision, to simplify the necessity of complex depth maps and temporal information. Their model creates a 14x14 score map which helps to perform pixel-wise binary supervision. On top of this supervision, their model is guided with binary cross-entropy. We build our model on top of \cite{deeppixbis} by using their pixel-wise binary supervision and model architecture. Instead of using binary cross-entropy, we propose to use only pixel-wise binary supervision. Moreover, we train our models with shuffled face images that are created by multiple patches of different face parts of different subjects.

\cite{patch_depth_cnn} proposes a two-stream CNN that uses patches and depth maps. They claim that patch-based CNN learns to discriminate artefact patches independent of the spatial face areas whereas depth-based CNN allows the model to learn how a face-like holistic depth map should look like.  Our work, which combines pixel-wise binary supervision with patch-based CNN, is inspired by \cite{deeppixbis, patch_depth_cnn}. As \cite{patch_depth_cnn} showed, we believe that training a CNN with face patches allows the model to distinguish artefacts without learning background or dataset-specific traces. Therefore, it prevents the model to overfit. Effectiveness of similar patch  shuffling method  is presented in \cite{patchshuffle}. In \cite{patchshuffle} it is shown that shuffling pixels within a patch can increase the generalization of the models. Our approach tries to mimic this behaviour from a higher level of view.  

Currently, the state-of-the-art frame-level PAD method \cite{cdcn} leverages a novel convolution operation. Authors propose Central Difference Convolution (CDC) for detecting detailed artefact traces. CDC specifically focuses on artefact traces. Their model is trained with pseudo-depth maps which require additional pseudo-depth map creation steps for ground truth. They also use computationally expensive Neural Architecture Search (NAS) to find a better and more efficient model which they call CDCN++. They report the lowest error rates on the OULU-NPU \cite{OULU_NPU_2017} dataset.

  \begin{figure}[t]
	\centering
	\includegraphics[width=\linewidth]{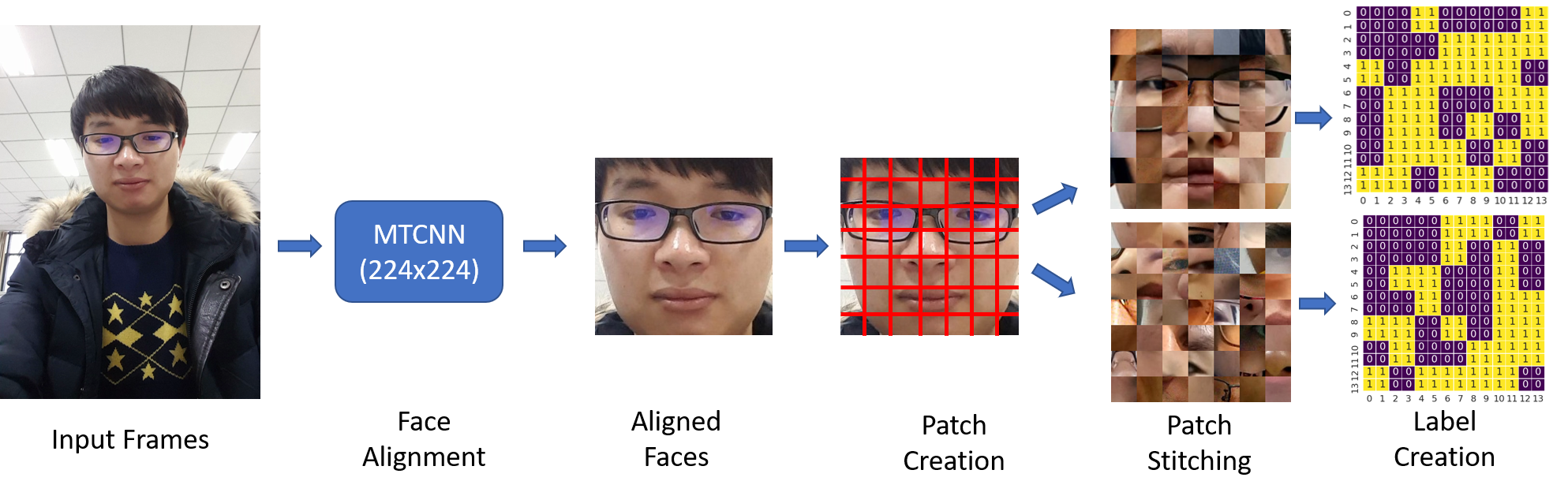}
	\caption{\label{fig:patchcreation} Our overall pipeline of input and ground truth label creation.}
\end{figure}

\section{Methodology}
%\subsection{Creation of Inputs and Labels }
Preprocessing of the images is an important task in PAD systems. The preprocessing pipelines are very similar among different methods. We first detect faces and face landmarks in the given frame with MTCNN~\cite{mtcnn} face detector. Then by using Bob framework \cite{bob2012,bob2017} we align the detected faces according to the eye coordinates and crop these aligned faces in 224x224 resolution. After that, we create patches from each face image by dividing the face into 7x7. Therefore, we get 49 face patches with 32x32 resolutions for each face image. We then combine these patches to create a new 224x224 image. Each face patch corresponds to a 2x2 location in the ground truth 14x14 label map. Bona fide patches have 1 as the label and fake patches have 0 as the label. We call this process patch stitching. We use two different strategies while stitching the patches. In the first approach, which we call random stitching, we completely randomize the patches and do not care about facial structure while combining different patches. Therefore, in this method, we get completely shuffled face images. As we randomly choose patches, the same parts of faces can be found in stitched images. For example, multiple noses or eyes can be seen in the bottom part of the \autoref{fig:patchcreation}. In the second approach, which we call controlled-stitching, we combine patches of different faces while keeping the facial structure as much as possible. Therefore, we actually create an input that resembles a face and consists of multiple subjects' face parts. The input contains both bona fide and attack patches, therefore stitched images must have labels for each patch separately. In our experiments we use the former approach which gives better performance in inter-dataset experiments. The overview of the proposed method is illustrated in the \autoref{fig:arch-illustration}.  

\subsection{Model Architecture and Training}
We use a deep CNN network that takes 224x224 input images and creates a 14x14 output map. We use the model that was proposed in \cite{deeppixbis} which is based on DenseNet-161 \cite{densenet} architecture. However, we do not have the final linear and sigmoid layers. The model contains the first eight layers of the DenseNet \cite{densenet} and we use pretrained weights. At the end of eight layers, we add a 1x1 convolution to produce a 14x14x1 feature map which is the model output.%We also use the first 6 stages of the EfficientNet-B7\cite{efficientnet} which takes 224x224 input and produces 14x14x256 feature maps. Then we use 1x1 convolutions to produce a 14x14x1 feature map. %We use the EfficientNet model to show that our proposed training method is not specific to architecture and can be used with different backbones.  
 We use input images that contain multiple patches together, therefore, we do not have a binary label. In training time, we use 14x14 pixel-wise binary labels to train the model with Binary Cross Entropy (BCE) loss. We assign ground truth y=0 for patches that come from attack images and y=1 for patches that come from bona fide images. The equation for pixel-wise loss is shown below. 

\begin{equation}
\label{eq:loss}
\mathcal{L}_{pixel-wise BCE} = -(y(log(p)) + (1-y)log(1-p)) 
\end{equation}

In \autoref{eq:loss} \textit{p} is the  14x14 model output that contains probability values between 0 and 1.  We minimize this loss with Adam \cite{adam} optimizer. We use 0.001 as the initial learning rate and halve this value at each 10th epoch. We use 32 as batch size and we generally train our methods for 30 epochs. We use horizontal flip and color jitter as data augmentation. In the test time, we give the aligned face image to the model and our model gives 14x14 output. We calculate the mean probability score by using the 14x14 output, then we use this probability as our liveness score. If the score is higher than the predefined threshold, we classify the given input as bona fide, else we classify it as an attack.

\begin{figure*}[t]
\centering
\subfloat[Samples from Sodec Real-World dataset][Samples from Sodec Real-World dataset]{
\includegraphics[width=0.46\textwidth]{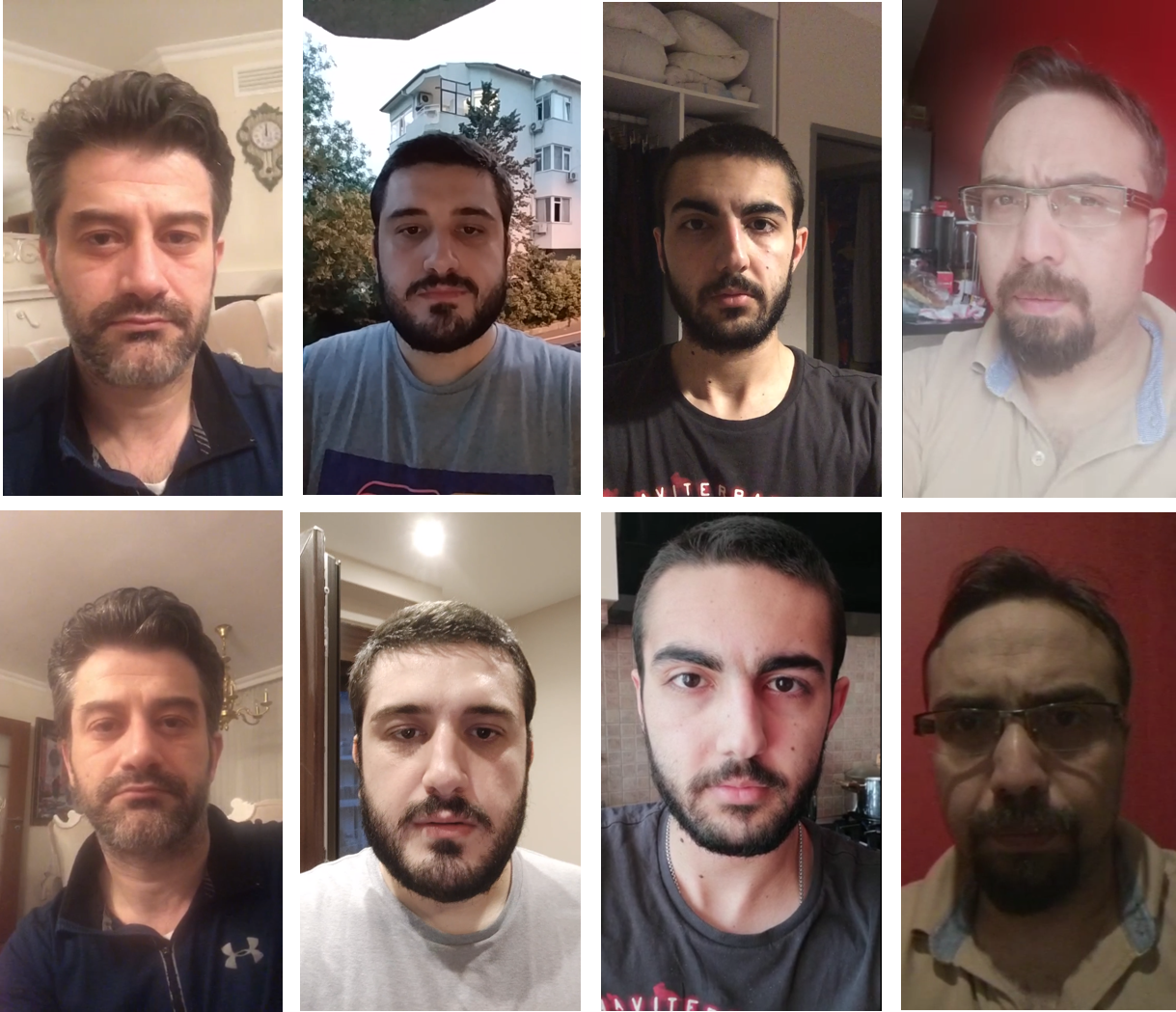}
\label{fig:subfig1}}
\qquad
\subfloat[Samples from OULU-NPU dataset][Samples from OULU-NPU dataset]{
\includegraphics[width=0.46\textwidth]{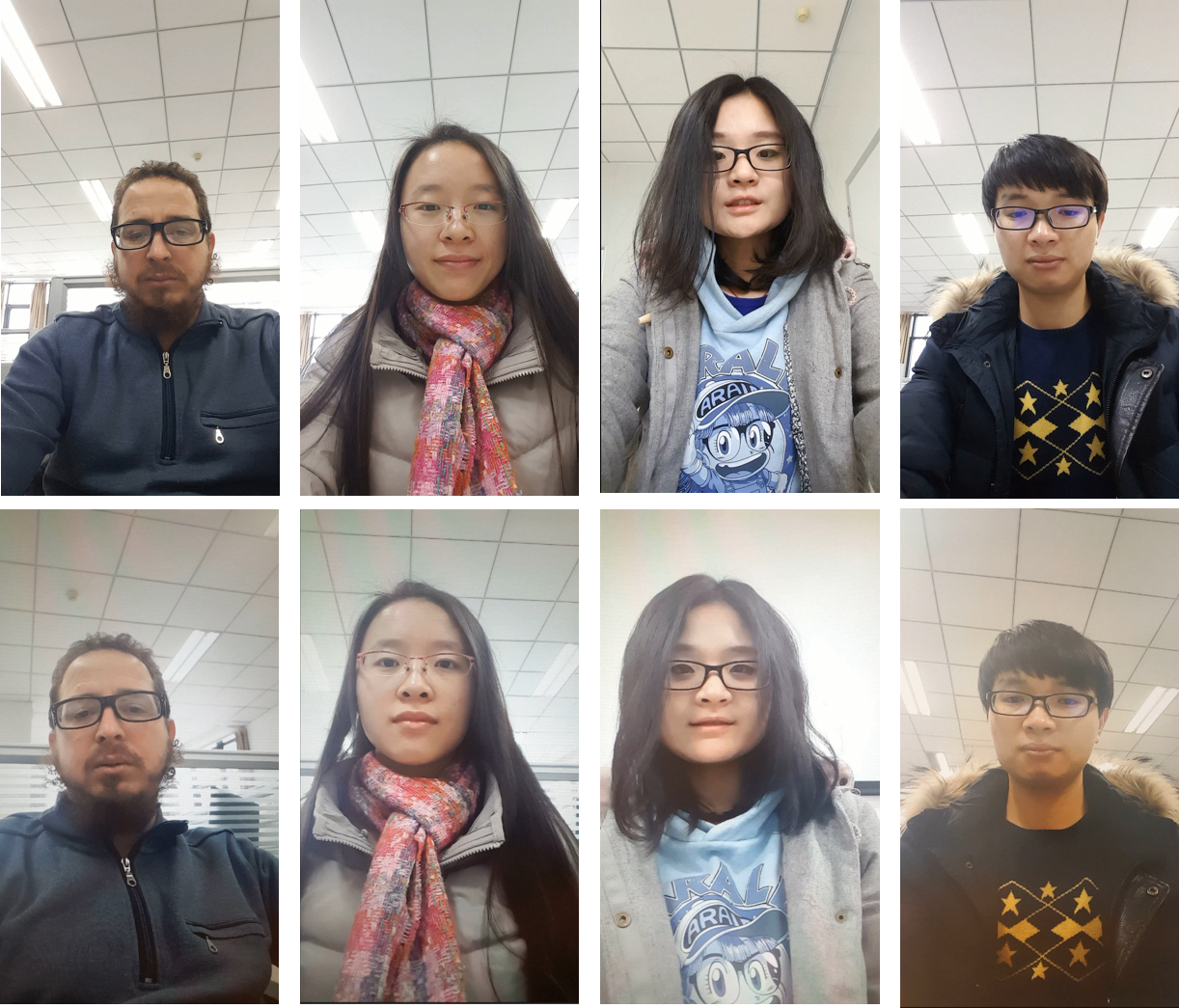}
\label{fig:subfig2}}
\caption{Sample images from OULU-NPU and Sodec Real-World datasets.}
\label{fig:oulu_realworld}
\end{figure*}

\section{Experimental Results}
This section presents the datasets that have been used to assess the performance of the proposed approach, the experiments carried out, and the obtained results.
\subsection{Datasets and Metrics} 
\textbf{Replay Mobile:} Replay Mobile dataset \cite{replaymobile} consists of 1190 video clips of 40 subjects. It contains paper and replay presentation attacks under five different lighting conditions. Each subject has 10 videos under different lighting and background conditions. %All videos of the dataset are at least 10 seconds long and captured with a 720p back camera of iPad Mini2 (running iOS) and an LG-G4 smartphone (running Android). 
There are mainly two kinds of attacks; matte screen and print. Matte screen attacks are the replay attack scenario where subject videos are displayed on a 1080p monitor, then recorded off of it. %For all matte screens, a stand has been used to hold capturing device. 
In print attacks, the faces of the subjects are printed on A4 paper then put on a stationary surface. %For print attacks, if capturing device has been held by a spoofer it is considered as the "hand" category or if it has been held by a stand then it is considered as the "fixed" category. 
The dataset has \textit{"grandtest"} protocol for evaluating the global performance of an algorithm. We report our results on the \textit{"grandtest"} protocol. %We use the train, devel, and test sets of the dataset provided with the dataset for training and evaluating our models. 

\textbf{OULU-NPU:} OULU-NPU dataset \cite{OULU_NPU_2017} is a high resolution antispoofing dataset. It has over 5900 videos of 55 subjects. %Videos were recorded in three different environments to ensure the variety of the real world. All of the videos have a 1080p resolution and are taken by 6 different mobile phone cameras. 
The dataset has both print and replay attacks with two printers and two display devices. There are 4 protocols for evaluating the methods' generalization capabilities. The protocol names were set based on the level of difficulty.  %In all of these protocols subjects of train, development, and testing sets are unique. Therefore in the test set, there are people that our algorithm has never seen before. Protocol 1 evaluates the generalization of the face PAD methods under previously unseen background and illumination conditions. The first two sessions are used for training and the last session is used for testing. Protocol 2 evaluates the generalization across different attack devices such as different printers or displays. Protocol 3 uses recording from different phones to test the method's generalizability across different input sensors. Protocol 4 is the most challenging protocol which evaluates the method across previously unseen environmental conditions, attacks and input sensors. It combines every restriction in the previous protocols. Moreover, it contains less training data. We report mean and standard deviation metrics on Protocols 3 and 4 which use the leave-one-out strategy for 6 different phone cameras.

\textbf{SiW:} SiW dataset \cite{siwdataset} is one of the largest high-quality antispoofing datasets. It has over 4400 videos of 165 subjects collected over 4 different sessions. All videos have 1080p resolution and contain variations of distance, pose, illumination, and expression. It contains both print and replay attacks. The dataset has 3 protocols to test the generalization of the models. We utilize this dataset for inter-dataset experiments because of the following reasons: it contains one of the highest numbers of subjects among PAD datasets, it contains different poses and expressions, and the acquisition device changes its distance to subjects which is a common behavior in the real world PAD attempts.  

\textbf{Sodec Real-World Dataset:} Sodec Real-World dataset has been collected to simulate the real-world presentation attack scenarios. It contains more than 51k frames of 31 different subjects. There are 16 male and 15 female subjects. It contains only replay attacks over 3 different presentation attack instruments (PAI), namely mobile phone, computer display, and television. Unlike controlled datasets, every subject recorded real videos with their mobile phone in their own home. Moreover, subjects were asked to rotate the phone vertically on the spot themselves while holding the phone in their outstretched arms and recording the video of themselves. It allows us to capture different backgrounds, illumination conditions, and pose variations with different input sensors. We show some examples from the dataset in \autoref{fig:oulu_realworld}. We utilize this dataset in inter-dataset experiments to test models on real-world attacks. %We divide the dataset into 21 and 10 subjects as train and test sets, respectively. 

%\subsection{Metrics}
In all of our experiments, we used  ISO/IEC 30107-3 \cite{iso30107} standards which are standard metrics for the PAD. Attack  Presentation Classification Error Rate (APCER), Bona fide Presentation Classification Error Rate (BPCER), Half Total Error Rate (HTER) along with the Average Classification Error Rate (ACER) in the test set are reported in the experiments. For all of our experiments, we use the threshold according to the equal error rate (EER) criterion. Average of False Recognition Rate (FRR) and False Acceptance Rate (FAR) is equal to HTER. We show the calculation of these metrics in \autoref{eq:metrics1} and \autoref{eq:metrics2}. 

\begin{equation}
ACER = \frac{APCER + BPCER}{2} 
\label{eq:metrics1}
\end{equation}
\begin{equation}
HTER = \frac{FRR + FAR}{2} 
\label{eq:metrics2}
\end{equation}

\begin{table}[b]
\centering
\begin{tabular}{|l|c|c|} 
\hline
\textbf{Model}               & \multicolumn{1}{l|}{\textbf{EER(\%)}} & \multicolumn{1}{l|}{\textbf{HTER(\%)}}  \\ 
\hline
\textbf{CDCN}          & \textbf{0.0}                          & \textbf{0.0}                            \\
\textbf{DeepPixBis}          & \textbf{0.0}                          & \textbf{0.0}                            \\
\textbf{Ours}     & \textbf{0.0}                          & \textbf{0.0}                            \\

\hline
\end{tabular}
\caption{Intra-dataset test results of Replay-Mobile \textit{"grandtest"} protocol. The first column represents the Equal Error Rate (EER) in percentage and the second represents the Half Total Error Rate (HTER) in percentage. }
\label{tab:replay}
\end{table}
%For cross-database testing, Half Total Error Rate (HTER) is used following the previous work in the literature. HTER is calculated by taking the average of False Rejection Rate (FRR) and the False Acceptance Rate (FAR).
\subsection{Experiments and Results}
We use OULU-NPU and Replay Mobile datasets for our intra-dataset experiments. As explained above, on OULU-NPU we report APCER, BPCER, and ACER performances of our and other models. On Replay Mobile dataset we report EER and HTER performances. We compare our method with CDCN++ \cite{cdcn} and DeepPixBis \cite{deeppixbis} models. CDCN++ is a state-of-the-art method that uses pseudo-depth maps and NAS. DeepPixBis has the same CNN architecture as our model. We differ from DeepPixBis in only our input and label creation where we utilize patch-wise labels. Therefore, our proposed method is directly comparable with DeepPixBis \cite{deeppixbis}. Similar to us, in the Replay-Mobile dataset, most of the newest methods report 0\% error. \autoref{tab:replay} shows that our proposed method also achieves 0\% error on the dataset. In \autoref{tab:oulu} we report our intra-dataset results on the OULU-NPU dataset. \autoref{tab:oulu} shows that our method outperforms DeepPixBis on Protocol 3 and Protocol 4 which are the hardest protocols in the dataset. These protocols have smaller training data and their test data is not very similar to the training data in terms of environment, PAI, PA acquisition device. Our method has a higher error rate on Protocol 1 and Protocol 2, but according to the our experiment results we show that our method is well suited for generalization whereas other methods gain an advantage of similar training and testing sets in these protocols. Our result does not outperform the state-of-the-art PAD model, however, CDCN and CDCN++ use pseudo-depth maps. In the case of CDCN++, it employs computationally expensive NAS operations.

\begin{table}[t]
\centering
\begin{tabular}{|c|c|c|c|c|} 
\hline
\textbf{Protocol}                        & \textbf{Model}       & \textbf{APCER(\%)}   & \textbf{BPCER(\%)}   & \textbf{ACER(\%)}     \\ 
\hline
\multicolumn{1}{|c|}{\multirow{4}{*}{1}} & CDCN        & \textbf{0.4}         & 1.7                  & 1.0                   \\
\multicolumn{1}{|c|}{}                   & CDCN++      & \textbf{0.4}         & \textbf{0.0}         & \textbf{0.2}          \\
\multicolumn{1}{|c|}{}                   & DeepPixBis  & \underline{0.83}                 & \underline{\textbf{0.0}}         & \underline{0.42}                  \\
\multicolumn{1}{|c|}{}                   & \textbf{Ours}       & 2.14                 & 2.14                 & 2.14      \\               
% \\ \multicolumn{1}{|c|}{} & Ours(EfficientNet)   & 2.54 & 2.58 & 2.56 \\ 
\hline
\multicolumn{1}{|c|}{\multirow{4}{*}{2}} 
                                         & CDCN        & \textbf{1.5}         & 1.4                  & 1.5                   \\
                                         & CDCN++      & 1.8                  & \textbf{0.8}         & \textbf{1.3}          \\
                                         & DeepPixBis           & 11.39                & \underline{0.56}                 & \underline{5.97}                  \\
                                         & \textbf{Ours}       & \underline{6.22}                 & 6.26                 & 6.24\\ 
                                         %& Ours(EfficientNet)   & 5.10 & 5.89 & 5.49 \\ 
\hline
\multicolumn{1}{|c|}{\multirow{4}{*}{3}} 
                                         & CDCN                 & $2.4\pm1.3$          & $2.2\pm2.0$          & $2.3\pm1.4$           \\
                                         & CDCN++      & $\textbf{1.7}\pm\textbf{1.5}$ & $\textbf{2.0}\pm\textbf{1.2}$ & $\textbf{1.8}\pm\textbf{0.7}$  \\
                                         & DeepPixBis           & $11.67\pm19.57$      & $10.56\pm14.06$      & $11.11\pm9.4$         \\
                                         & \textbf{Ours}       & $\underline{6.10}\pm2.57$        & $\underline{6.30}\pm2.55$        & $\underline{6.20}\pm2.55$                 \\% & Ours(EfficientNet)   & $\pm$ & $\pm$ & $\pm$ \\ 
\hline
\multicolumn{1}{|c|}{\multirow{4}{*}{4}} 
                                         & CDCN                 & $4.6\pm4.6$          & $9.2\pm8.0$          & $6.9\pm2.9$           \\
                                         & CDCN++      & $\textbf{4.2}\pm\textbf{3.4}$ & $\textbf{5.8}\pm\textbf{4.9}$ & $\textbf{5.0}\pm\textbf{2.9}$  \\
                                         & DeepPixBis           & $36.67\pm29.67$      & $13.33\pm16.75$      & $25.0\pm12.0$         \\
                                         & \textbf{Ours}       & $\underline{11.51}\pm7.86$                & $\underline{11.58}\pm7.84$                & $\underline{11.54}\pm7.84$ \\ 
% & Ours(EfficientNet)   & $\pm$ & $\pm$ & $\pm$ \\ 
\hline
\end{tabular}
\caption{Intra-dataset test results of OULU-NPU dataset.}
\label{tab:oulu}
\end{table}

\subsection{Real-World Experiments}

There are many face antispoofing datasets that are collected in controlled environments. Most of these datasets only consider two or three backgrounds with controlled illumination changes. However, in real-world applications, there are various backgrounds, illumination, poses, and expression changes. Moreover, attackers are more careful when creating a face presentation attack. We have collected a dataset to simulate the real-world use case of the presentation attacks. We trained CDCN \cite{cdcn}, DeepPixBis \cite{deeppixbis}, and our model on the SiW dataset Protocol 1. %SiW dataset is also captured in a controlled environment, however, 
We choose the SiW dataset because it is one of the most representative datasets in terms of distance, pose, illumination, and expression changes. From \autoref{tab:crossdataset} we see that state-of-the art CDCN model has achieved the lowest error rate in the SiW dataset, but gets a higher error rate on Sodec Real-World Dataset. Our method is able to outperform DeepPixBis on both the SiW dataset and Sodec Real-World Dataset. The results show that our proposed training method is useful for real-world inter-dataset scenarios which is the hardest task to perform. 

\begin{table}[t]
\centering
\begin{tabular}{|l|c|c|c|c|} 
\hline
                             & \multicolumn{2}{c|}{\textbf{Trained on SiW}}                             \\ 
\hline
\textbf{Model}               & \multicolumn{1}{l|}{tested on SiW} & \multicolumn{1}{l|}{tested on Sodec Real-World Dataset}  \\ 
\hline
CDCN                         & \textbf{0.12} & 12.52          \\
DeepPixBis                   & 3.68 & 12.05    \\
\textbf{Ours}     & 2.15 & \textbf{5.24}  \\
%\textbf{Ours (EfficientNet)} & 1.97 & \textbf{4.06}   \\
\hline
\end{tabular}
\caption{Experiment results of SiW (Protocol 1) and inter-dataset test results on Sodec Real-World Dataset. Reported metrics in the table represents the ACER values in percentage (\%)}
\label{tab:crossdataset}
\end{table}

\section{Conclusion and Future Work}

In this paper, we propose a new training method for the PAD models. Our proposed method uses combined face patches instead of one single face image in training time. We show that training models with pixel-wise binary loss and shuffled face patches can improve PAD performance. Our proposed method improves DeepPixBis' \cite{deeppixbis} performance on OULU-NPU Protocol 3 and 4 which are the hardest protocols. Moreover, the proposed method performs much better when we test the models on a real-world dataset. For future work, we are planning to extend the Sodec Real World dataset to print attacks. Furthermore, we are planning to modify the backbone architecture of the model and analyze the effects of different patch creation methods on model behavior.

\section{Acknowledgements}

This work was partially supported by a Sodec Technologies research grant. We would like to thank Sodec Technologies for their data collection efforts and support for this work. 

\bibliography{references}

\end{document}